\documentclass[acmsmall]{acmart}
\usepackage{multirow}
\usepackage{graphicx}
\usepackage{float}                  % 图片浮动位置
\usepackage{subfig}                 % 子图包，不要与{subfigure}混用
\usepackage{overpic}                % 与图片排版相关
\usepackage{makecell} 
\usepackage{afterpage}

\AtBeginDocument{%
  }

\setcopyright{acmcopyright}
\copyrightyear{2018}
\acmYear{2018}
\acmDOI{XXXXXXX.XXXXXXX}

\acmJournal{JACM}
\acmVolume{37}
\acmNumber{4}
\acmArticle{111}
\acmMonth{8}

\begin{document}

\title{SigFormer: Sparse Signal-Guided Transformer for Multi-Modal Action Segmentation}

%% "authornote" and "authornotemark" commands \authornotemark[1]
\author{Qi Liu}
\affiliation{%
  \institution{Institute of Information Engineering, Chinese Academy of Sciences \& School of Cyber Security, University of Chinese Academy of Sciences \&
  Key Laboratory of Cyberspace Security Defense}
  \streetaddress{No. 19 Shucun Road, Haidian District}
  \city{Beijing}
  \state{Beijing}
  \country{China}
  \postcode{100085}
}
\email{liuqi@iie.ac.cn} 

\author{Xinchen Liu}
\affiliation{%
  \institution{JD Explore Academy}
  \streetaddress{Courtyard No. 18, Kechuang 11th Street, Daxing District}
  \city{Beijing}
  \state{Beijing}
  \country{China}
  \postcode{100176}}
\email{xinchenliu@bupt.cn}

\author{Kun Liu}
\authornote{Corresponding author.}
\affiliation{%
  \institution{JD.com Inc}
  \streetaddress{Courtyard No. 18, Kechuang 11th Street, Daxing District}
  \city{Beijing}
  \state{Beijing}
  \country{China}
  \postcode{100176}
}
\email{liu_kun@bupt.cn}

\author{Xiaoyan Gu}
\authornotemark[1]
\affiliation{%
  \institution{Institute of Information Engineering, Chinese Academy of Sciences \& School of Cyber Security, University of Chinese Academy of Sciences
  }
  \streetaddress{No. 19 Shucun Road, Haidian District}
  \city{Beijing}
  \state{Beijing}
  \country{China}}
  \postcode{100085}
\email{guxiaoyan@iie.ac.cn}

\author{Wu Liu}
\affiliation{%
  \institution{School of Information Science and Technology, University of Science and Technology of China}
  %\streetaddress{Courtyard No. 18, Kechuang 11th Street, Daxing District}
  \city{Hefei}
  \state{Anhui}
  \country{China}
  \postcode{230022}}
\email{liuwu@ustc.edu.cn}

\renewcommand{\shortauthors}{Q. Liu et al.}


\begin{abstract}

Multi-modal human action segmentation is a critical and challenging task with a wide range of applications. 
Nowadays, the majority of approaches concentrate on the fusion of dense signals (i.e., RGB, optical flow, and depth maps). 
However, the potential contributions of sparse IoT sensor signals, which can be crucial for achieving accurate recognition, have not been fully explored. 
To make up for this, we introduce a \textbf{S}parse s\textbf{i}gnal-\textbf{g}uided Transformer (\textbf{SigFormer}) to combine both dense and sparse signals. 
We employ mask attention to fuse localized features by constraining cross-attention within the regions where sparse signals are valid. 
However, since sparse signals are discrete, they lack sufficient information about the temporal action boundaries.
Therefore, in SigFormer, 
we propose to emphasize the boundary information at two stages to alleviate this problem. 
In the first feature extraction stage, we introduce an intermediate bottleneck module to jointly learn both category and boundary features of each dense modality through the inner loss functions. 
After the fusion of dense modalities and sparse signals, 
we then devise a two-branch architecture that explicitly models the interrelationship between action category and temporal boundary. 
Experimental results demonstrate that SigFormer outperforms the state-of-the-art approaches on a multi-modal action segmentation dataset from real industrial environments, 
reaching an outstanding F1 score of 0.958. 
The codes and pre-trained models have been available at https://github.com/LIUQI-creat/SigFormer. 
\end{abstract}




% \received[revised]{12 March 2023}
% \received[accepted]{5 June 2023}

\maketitle

\section{INTRODUCTION}
Multi-modal action segmentation is a fundamental task in the multi-media community, 
which aims to classify action labels in time-series data at the frame level using multiple signals~\cite{lea2017temporal, kuehne2014language, yi2021asformer, wang2020boundary, farha2019ms, yoshimura2022openpack}. 
This task is extremely challenging since it is required to combine multiple data sources to learn the semantics of fine-grained action categories and corresponding temporal boundaries. 
Consequently, multi-modal action segmentation has sparked substantial research interest and has been widely applied in various domains, 
including human behavior analysis~\cite{zeng2021graph, vallacher1987people, huang2020improving}, 
video surveillance~\cite{collins2000introduction, collins2000system, liu2019exploring}, 
and worker action understanding in industrial fields~\cite{yoshimura2022openpack, niemann2020lara}.

\afterpage{
\clearpage
\begin{figure}[h]
    \centering
    \includegraphics[width = \linewidth]{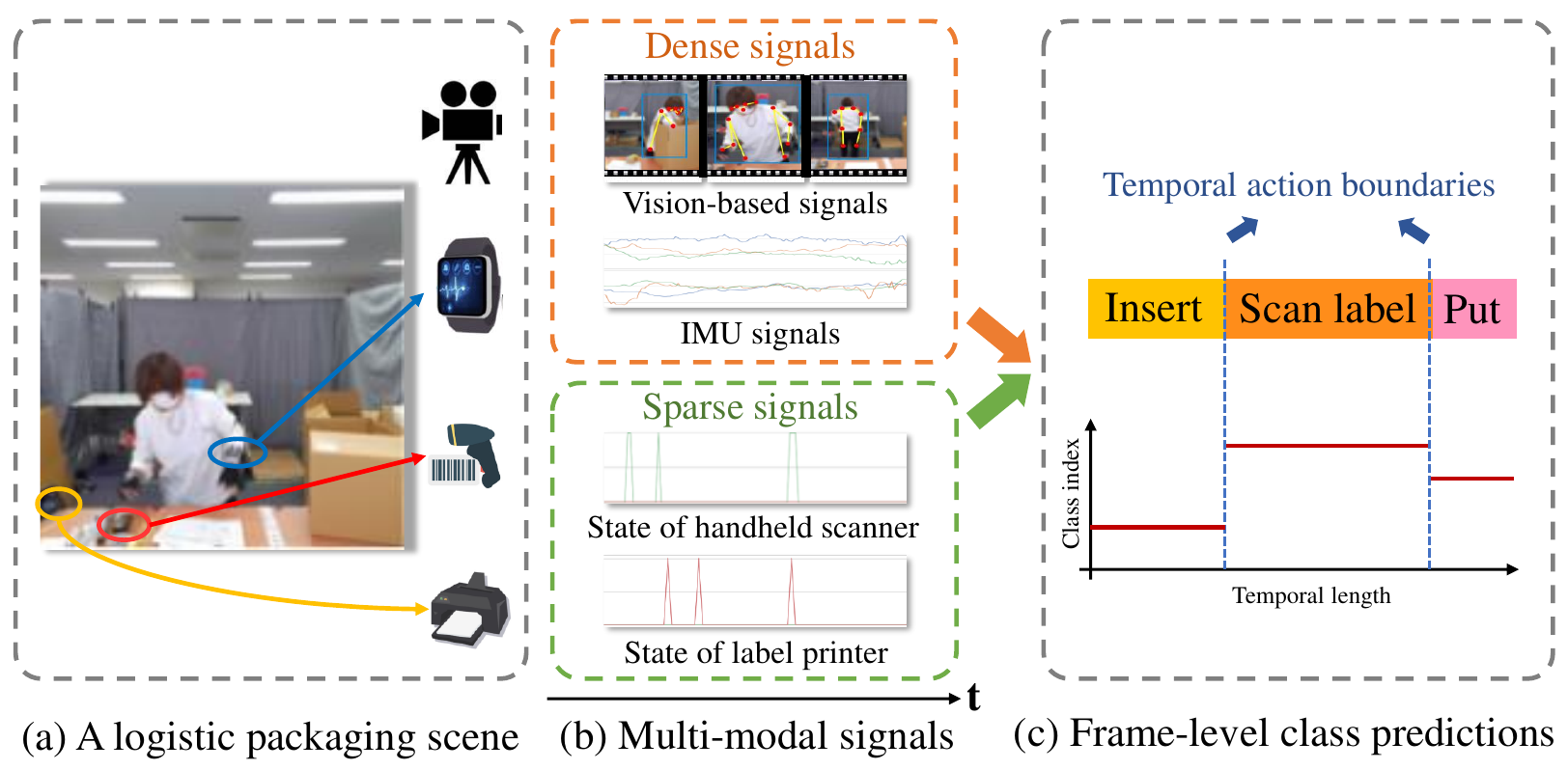}
    \caption{
    Example of action segmentation utilizing multi-modal signals.
    (a) A worker is engaged in a packaging task within a scenario equipped with various devices and sensors, including cameras, inertial measurement units (IMUs), and Internet of Things (IoT) devices like handheld scanners and label printers. 
    (b) These components generate diverse multi-modal signals, which include both dense signals, such as vision-based data representing human key points and bounding boxes, as well as sparse signals derived from the state of IoT devices.
    (c) We employ these multi-modal signals to segment the worker's actions temporally to generate frame-level class predictions, which in detail means predicting the action category at a frequency of 1 Hz.
    Given that the input data is untrimmed, action segmentation necessitates not only categorizing the action class but also precisely identifying the temporal boundaries between actions.
    }
    \label{fig:fig1}
    \Description{Example of action segmentation utilizing multi-modal signals.  It shows that a worker is engaged in a packaging task within a scenario equipped with various devices and sensors, including cameras, inertial measurement units (IMUs), and networked devices like handheld scanners and label printers.}
\end{figure}

}

Although several multi-modal action segmentation methods have achieved promising performance~\cite{farha2019ms, wang2020boundary, ishikawa2021alleviating, chen2020action, gao2021global2local, park2022maximization, chen2022uncertainty, behrmann2022unified, liu2023diffusion, yi2021asformer}, they exclusively employ dense signals as input to predict the action category. 
For example, RGB, audio, optical flow, depth, and skeleton are usually chosen to combine for multi-modal action segmentation. 
In contrast, few studies classify the activities with sparse signals, such as the states of Internet of Things (IoT) devices which are closely associated with human activities. 
The signals originating from IoT devices (i.e., printer and scanner) are temporally sparse because they are primarily recorded when a person operates these devices, with they spending substantial durations in an inactive state.
As a result, they have ideal properties to aid accurate recognition in some categories. 
As shown in Fig.~\ref{fig:fig1}, in the logistic packaging scene, the states of printers and scanners can be beneficial to selecting the key moments that are useful for recognizing the "scan label" category. 
Generally, the logs of sensors are extremely clean and contain no noise, 
which plays a pivotal role in achieving accurate identification. 
However, fusing the sparse and dense data sources for action segmentation still faces two challenges.

First of all, in the field of multi-modal signal fusion, a fundamental challenge arises from the inherent disparities between sparse and dense signals. 
Sparse signals provide limited yet precise information, while dense signals contain abundant information but are susceptible to noise and errors.
The variance in information density between these signal types makes it challenging to balance the importance and weight between them during the fusion process.
Additionally, when it comes to the accuracy of action representation, how to guide continuous dense signals with precise sparse data points to minimize noise interference presents a problem that needs to be solved.

Another aspect concerns the ability to be aware of the temporal action boundaries. 
While the sparse IoT sensor signals possess favorable characteristics for recognizing certain categories, they lack adequate awareness of temporal action boundaries which are important for action segmentation tasks.
Dense predictions of action segmentation, commonly trained with frame-wise supervision, can easily cause over-segmentation~\cite{wang2020boundary, ishikawa2021alleviating}. 
It means that the occurrence of short misclassifications within a long action segment, which seriously affects the performance of action segmentation. 
One potential reason is that the slight variances between consecutive frames cause boundary ambiguity~\cite{wang2020boundary}, leading to the network’s failure to converge on such semantically uncertain boundaries.
Consequently, developing effective strategies for integrating sparse and dense signals becomes essential.
Furthermore, It is also critical to enhance the network’s awareness of action temporal boundaries.

Many methods have been proposed~\cite{tomoon, yoshimura2022openpack, inoshita2023exploring} for the task of multi-modal action segmentation. 
For instance, Uchiyama~\cite{tomoon} merely concatenated multi-modality features with different information densities and fed them into a joint feed-forward network. 
The feature concatenation method employed here overlooks the need to balance the weights assigned to each signal type and fails to leverage their individual strengths effectively.
Yoshimura {\itshape et al.}~\cite{yoshimura2022openpack} leveraged the high-confidence sparse signals as anchors and utilized bidirectional feedback blocks to fuse these anchors with the dense sensor signals. 
While this kind of method generates anchors to utilize accurate cues in sparse signals, it does not adapt to variations in the data and ignores temporal action boundary information. 
Inoshita {\itshape et al.}~\cite{inoshita2023exploring} utilized 1D convolutional layers to cross-weight two-by-two feature sets of multi-modal signals in a mutual cross fusion module. 
This combination of fixed pairs of modalities may not scale well when dealing with a large number of modalities.
Furthermore, it's worth noting that certain combinations of modalities lack sparse signals, and even with the presence of accurate cues within the sparse signals, they may not entirely mitigate the noise generated by other signals.
In conclusion, these methods do not completely solve the problems in the process of fusing sparse and dense signals. 

To tackle the challenge of multi-modal action segmentation utilizing sparse and dense signals, we propose a \textbf{S}parse s\textbf{i}gnal-\textbf{g}uided Transformer (\textbf{SigFormer}). 
Our objective is to explore an effective method to utilize multi-modal signals to attain accurate representations of fine-grained actions.
In SigFormer, we propose a sparse guided cross-modal module designed to integrate multi-modal signals with various information densities, aiming to achieve a comprehensive representation of human actions.
It consists of two modules: \textbf{S}parse \textbf{G}uided \textbf{F}usion (\textbf{SGF}) module and \textbf{M}otion-\textbf{S}patial \textbf{A}ttention \textbf{F}usion (\textbf{MSAF}) module. 
The SGF module is responsible for fusing sparse state signals from IoT-enabled devices with dense signals.
Within the SGF module, we utilize masked attention to restrict the attention to the localized feature fusion within temporal segments where sparse signals are valid.
Compared to other fusion methods and the typical cross-attention used in a standard Transformer, our SGF module with masked attention leads to superior performance. 
The MSAF module is tasked with encoding the interaction context between motion and spatial information that has been precisely guided by sparse signals.
Furthermore, we emphasize the awareness of temporal action boundaries at two stages.
First, we introduce an intermediate bottleneck module to acquire task-specific aligned features for action segmentation.
In this module, instead of concentrating solely on action categories~\cite{tomoon, carreira2017quo, li2022bridge}, we enhance the distinctive features of each modality, encompassing both action category and boundary information by incorporating inner losses into the end-to-end network. 
This ensures that the features of multiple modalities are aligned within a consistent semantic space, thereby enabling the extraction of valuable information related to action segmentation.
Additionally, we incorporate two mutual interactive branches after feature fusion, which explicitly model the multi-stage interaction between the category branch and the boundary branch to further enhance the boundary awareness of the overall network.
By placing great emphasis on temporal action boundaries, we effectively alleviate over-segmentation errors.
Evaluation results of SigFormer on a large-scale multi-modal benchmark from real industrial environments demonstrate that it can achieve better results compared to state-of-the-art methods. 

In conclusion, our main contributions are as follows:
\begin{itemize}
\item We present SigFormer, a novel multi-modal action segmentation network that effectively leverages both sparse and dense signals to achieve a comprehensive representation of fine-grained human actions. 

\item To enable multi-modal interactions, we introduce the sparse guided cross-modal module. 
Through masked attention, we restrict attention to localized feature fusion within the temporal regions where the sparse signals are valid.
This module greatly facilitates effective multi-modal fusion with different information densities.

\item To emphasize the temporal action boundary, we introduce an intermediate bottleneck module to jointly learn both the category and boundary of each dense modality through the inner loss functions.
Furthermore, we facilitate multi-stage interactions between action category and temporal boundary in mutual interactive branches to further augment the overall network’s boundary awareness.

\end{itemize}

SigFormer achieves state-of-the-art performance on the OpenPack dataset by fusing multiple modalities, attaining a remarkable F1 score of 0.958.
Through comprehensive experiments, we not only evaluate the contribution of different modalities to action segmentation, but also demonstrate the effectiveness of increasing the awareness of temporal action boundary.

The rest of this article is organized as follows, Section~\ref{sec:realted} explores the related work. Section~\ref{sec:method} presents a detailed description of SigFormer. Section~\ref{sec:experiment} describes our implementation details and experiment results. Section~\ref{sec:conclusion} provides the conclusion of our article.

\section{RELATED WORK} 
\label{sec:realted}
\textbf{Temporal Action Segmentation.} Temporal action segmentation is a distinctive task in video understanding that deals with complex videos spanning several minutes. To capture long-range dependencies between actions, various temporal models have been utilized in the literature, ranging from early recurrent neural networks~\cite{donahue2015long, yue2015beyond, singh2016multi, yeung2018every} to temporal convolutional networks~\cite{lea2017temporal, farha2019ms, wang2020gated, gao2021global2local, li2021efficient, park2022maximization, liu2018t, liu2020real}, graph neural networks~\cite{huang2020improving, zhang2022semantic2graph}. More recently, Transformers~\cite{yi2021asformer, aziere2022multistage, li2022bridge, wang2022cross} has attracted extensive attention due to the potential of the attention mechanism in capturing long-range temporal dependencies in video sequences. ASFormer~\cite{yi2021asformer}, for instance, employed a well-designed Transformer decoder to refine the initial predictions from the encoder. Br-Prompt~\cite{li2022bridge} resorted to multi-modal learning by utilizing text prompts as supervision. SEDT~\cite{kim2022stacked} preserved local information along with global information, by adding an encoder with self-attention before every decoder. In addition, various types of attention-based decoders~\cite{wang2022cross, behrmann2022unified} can be utilized to refine the predictions. 
However, the aforementioned methods are primarily vision-based, which can restrict the real-world applicability of action segmentation due to challenges posed by occlusion and background interference.
In this paper, we extract valuable features from several sensor signals and comprehensively fuse them to obtain a comprehensive representation of fine-grained actions.

To solve the issue of over-segmentation, the boundary-aware cascade network~\cite{wang2020boundary} introduced local barrier pooling to aggregate local predictions by leveraging semantic boundary information. Ishikawa {\itshape et al.}~\cite{ishikawa2021alleviating} 
divided temporal action segmentation into frame-wise action classification and action boundary regression, which refines frame-level hypotheses of action classes utilizing predicted action boundaries. Chen {\itshape et al.}~\cite{chen2022uncertainty} estimated the uncertainty arising from ambiguous boundaries employing Monte-Carlo sampling. In this paper, we introduce the inner boundary loss and establish multi-stage interactions in the class branch and boundary branch to alleviate over-segmentation errors.

\textbf{Multi-Modality Action Recognition.} Multi-modality action recognition techniques have gained significant attention due to the complementary information offered by different modalities, leading to improved classification performance through the fusion of multiple feature streams. Previous approaches mainly utilized decision-level fusion strategies~\cite{cai2021jolo, hu2015jointly, wu2016multi, zhang2018fusing, zhao2017two}, such as score fusion, or relied on simplistic feature-level fusion techniques. For example, Liu and Yuan~\cite{liu2018recognizing} concatenated skeleton and pose heatmap features. Noori {\itshape et al.}~\cite{noori2020human} explored several methods of fusion for multi-representations of data from sensors and proposed a CNN-based general architecture. Recent methods have focused on designing correlation modules for modality fusion~\cite{li2020sgm, joze2020mmtm, cai2021jolo, zhang2022semantic2graph}. Li {\itshape et al.}~\cite{li2020sgm} developed bilinear pooling to enable the skeleton feature to guide on RGB feature at the semantic feature level. Joze {\itshape et al.}~\cite{joze2020mmtm} introduced the fusion of multi-modality features across different spatial dimensions, enabling the re-calibration of channel-wise features within each CNN stream utilizing the knowledge from multiple modalities. Zhang {\itshape et al.}~\cite{zhang2022semantic2graph} turned the video action segmentation and recognition problem into node classification of graphs and utilized GNNs to learn multi-modal feature fusion. 

Recently, there has been a shift in research focus towards Transformers~\cite{nagrani2021attention, cvt, swin, wave-vit, contex, RVT, wang2022cross, xu2022exploiting, li2022bridge, van2022gesture, dual_vit, van2023aspnet, DBLP:conf/cicai/ZhaoLHBZL22, DBLP:journals/tnn/XuJB22}, which are versatile perceptual models capable of effectively handling diverse data sources with minimal modifications to the model structure. 
For example, Wu {\itshape et al.}~\cite{cvt} introduced convolutions into Vision Transformer (ViT) to improve ViT in performance and efficiency. 
Liu {\itshape et al.}~\cite{swin} proposed the Swin Transformer, a hierarchical Transformer with Shifted windows. 
%This architecture has now become the general backbone of computer vision.
Yao {\itshape et al.}~\cite{wave-vit} constructed a Wavelet ViT that enables self-attention learning with lossless down-sampling over keys/values. 
Li {\itshape et al.}~\cite{contex} designed a Contextual Transformer to fully capitalize on the contextual information among input keys to guide the learning of dynamic attention matrices.
Mao {\itshape et al.}~\cite{RVT} proposed Robust ViT which has advanced robustness and generalization ability.
To explore the holistic interactions among all patches, Yao {\itshape et al.}~\cite{dual_vit} proposed Dual-ViT that elegantly exploits the global semantics for self-attention learning.

Although the attention operation in Transformers provides a straightforward approach to integrating various signals, it fails to consider information redundancy and exhibits limited scalability when applied to longer temporal sequences. 
To address these challenges, Nagrani {\itshape et al.}~\cite{nagrani2021attention} proposed an attention bottleneck to control the flow of attention between tokens from different data sources, ensuring that each modality shares only essential information with the others. Van {\itshape et al.}~\cite{van2023aspnet} explicitly distinguished between modality-shared and private feature representations within a long-range action segmentation model, leveraging the bottleneck mechanism to maintain the separation of features in subsequent temporal processing layers. However, these methods do not ensure the quality of multi-modal features before fusion, which might result in the integration of irrelevant information and the introduction of interference. In this paper, we propose an intermediate bottleneck module to regulate the feature extraction process of diverse data and employ accurate sparse signals to guide dense signals.

\section{METHODOLOGY}
\label{sec:method}
This section provides a comprehensive description of SigFormer. 
SigFormer takes dense signals, including IMU data~(I), body key points~(K), body bounding box points~(B), as well as sparse state signals of IoT-enabled devices~(D), as input, and produces frame-level refined class predictions as output. 
We commence with an overview of SigFormer, subsequently delving into an intricate presentation of its four constituent modules.

\subsection{Overview}
\label{sec:overview}
\begin{figure}[t]
    \centering
    \includegraphics[width = \linewidth]{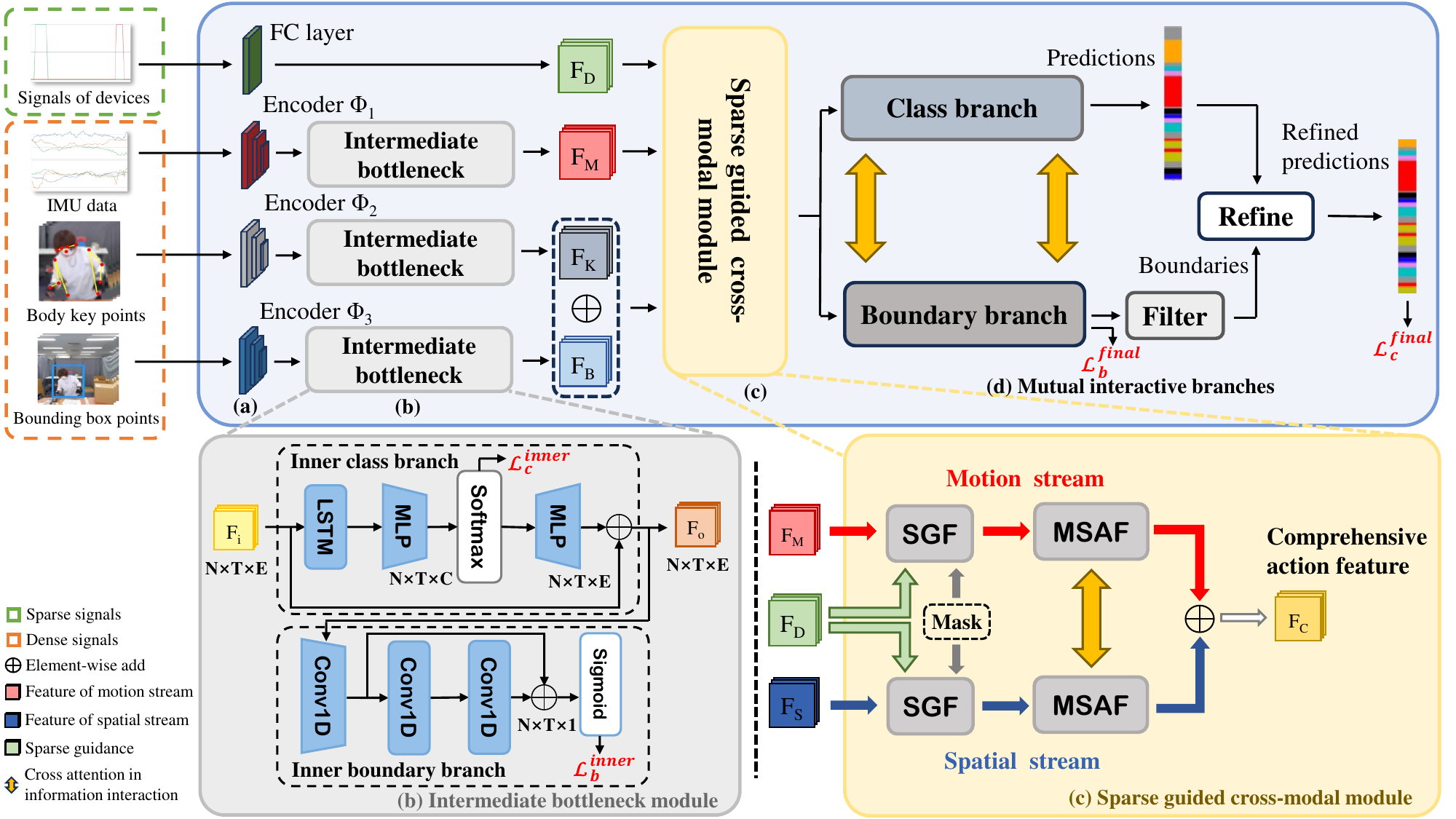}
    \caption{Overview of the pipeline for SigFormer. 
    (a) Feature extraction from each signal. % excluding RGB. 
    (b) The intermediate bottleneck module operates by taking $F_i$ ($i$ $\in M, K, B$) as input, reducing its feature dimension, calculating the inner class loss $\mathcal L_{c}^{inner}$, and the inner boundary loss $\mathcal L_{b}^{inner}$. 
    Subsequently, it restores the feature dimension to obtain $F_o$ ($o$ $\in M, K, B$), which possesses an enhanced awareness of both category and boundary.
    (c) The sparse guided cross-modal module combines features from multi-modal signals with different information densities, $F_M$, $F_D$, and $F_S$ (obtained by adding $F_K$ and $F_B$), to generate a comprehensive action representation $F_C$.
    This module comprises a sparse guided fusion (SGF) module and a motion-spatial attention fusion (MSAF) module. 
    (d) Mutual interactive branches employ Multi-Head Cross Attention (MHCA) to model information interaction between the class branch and boundary branch. The predictions of the class branch are refined by the boundaries (filtered from the predictions of the boundary branch) to obtain frame-level refined class predictions.}
    \label{fig:overall}
    \Description{An end-to-end network for solving multi-modality packing action segmentation tasks.}
\end{figure}

As depicted in Fig.~\ref{fig:overall}, the architecture of SigFormer comprises four components: the feature extraction module, the intermediate bottleneck module, the sparse guided cross-modal module, and the mutual interactive branches. 
Initially, feature extraction (Section~\ref{extrat}) is conducted on multi-modal data. 
Subsequently, the intermediate bottleneck module (Section~\ref{section:inner module}) is employed to impose constraints through the inner losses on both action class and temporal boundary for each modality.
This step provides pivotal and diverse information for subsequent modality fusion. 
To effectively leverage the complementary information across multiple modalities, the features are then partitioned into three categories: motion stream $F_M$ (the feature of IMU data), spatial stream $F_S$ (from the feature of body key points feature $F_K$ and the feature of bounding box points $F_B$) and sparse guidance $F_D$ (the feature of state signals of devices). 
Afterward, these features are fused in the sparse guided cross-modal module (Section~\ref{sec:fuse}) to attain a comprehensive representation of human actions~$F_C$.
Specifically, the sparse guidance undergoes integration with the motion and spatial stream features via the sparse guided fusion (SGF) module.
Within the SGF module, sparse signals accurately provide action class cues for dense signals through masked attention.
Following that, the motion-spatial attention fusion (MSAF) module implicitly formulates the correlation between the two streams 
Subsequently, the mutual interactive branches (Section~\ref{sec:fual-branch}) are employed to strengthen information interaction between action classes and temporal boundaries.
SigFormer identifies high-probability predictions from the boundary branch as temporal boundaries by a filter. 
Finally, these identified boundaries are employed to refine the results of the class branch, yielding frame-level refined class predictions.

\subsection{Feature Extraction Module} 
\label{extrat}
The IMU data includes acceleration data on three axes, as well as gyroscope and quaternion data, which is acquired by body-worn sensors. The human body key points and bounding box points are obtained from posewarper~\cite{pose} and faster r-cnn\cite{fasterrcnn}.
They offer noteworthy insights into motion and position at each temporal instant. 
Therefore, the demand arises for efficient techniques for feature extraction. 
To this end, we adopt the unified Transformer encoder-based layers~\cite{vaswani2017attention,tomoon} as the multi-modal feature extractors.

In contrast, IoT-enabled devices only get utilized during specific actions, making the state signals temporally sparse.
Consequently, a straightforward feature extractor proves adequate for this sparse data as
\begin{equation}
  F_D = W_D(D)
\end{equation}
where $D$ denotes the state signals of IoT-enabled devices, $W_D(\cdot)$ denotes a fully connected layer, and $F_D$ represents the resultant feature. 

Through the feature extraction module, we acquire the feature vector $F_x \in \mathbb{R}^{N\times T\times E_x}$ ($x \in M, K, B, D$), where $N$ represents the batch size, $T$ represents the length of the time series and $E$ represents the feature dimension.

\subsection{Intermediate Bottleneck Module}\label{section:inner module}
Most approaches~\cite{farha2019ms, ishikawa2021alleviating, yi2021asformer, li2022bridge} for action segmentation typically rely on pre-trained feature extractors~\cite{carreira2017quo, li2022bridge} for data processing and feature extraction, which are subsequently followed by feature fusion. 
However, these methods lack emphasis on capturing action boundary information within the data. 
Consequently, the resulting features inadequately localize the temporal boundaries between actions, thereby falling short of meeting the specific requirements of action segmentation. 
In contrast, we introduce a bottleneck module after the feature extraction of each modality, thus imposing additional constraints.
Through the enhancement of awareness regarding action class and temporal boundary, this module ensures that exclusively pertinent information essential for action segmentation is extracted from each modality.
As illustrated in Fig.~\ref{fig:overall} (b), the architecture of the intermediate bottleneck module contains two parts: the inner class branch and the inner boundary branch.

\textbf{Inner Class Branch.} It takes the feature $F_i \in \mathbb{R}^{N\times T\times E}$ ($i$ $\in M, K, B$) from the feature extraction module as input. 
Initially, the LSTM layer is employed to enhance the modeling capacity of the temporal dimension. 
Following that, an MLP layer $f_p(\cdot)$ is applied to reduce the feature dimension $E$ to fit the number of action classes $C$, and $softmax$ is employed to turn the prediction results into probabilities as
\begin{equation}
    \begin{split}
        & {F_{class}} = f_p(LSTM({F_{i}})), \\
	    & P_{class} = softmax({F_{class}})
    \end{split}
\end{equation}
where $F_{class}\in \mathbb{R}^{N\times T\times C}$ is the class output feature, $C$ represents the number of action classes ($C$ < $E$), $P_{class}$ is the predicted result of the inner class, from which the inner class loss $\mathcal L_{c}^{inner}$ is calculated. 
The feature $F_{class}\in \mathbb{R}^{N\times T\times C}$ is then fed into another MLP layer $f_p^{'}(\cdot)$, which adjustments the feature dimension back. 
Additionally, to prevent the potential loss of feature information resulting from intermediate dimensional reduction, we incorporate the input of the intermediate bottleneck module $F_i\in \mathbb{R}^{N\times T\times E}$ into the output of $f_p^{'}(\cdot)$ via a residual connection~\cite{he2016deep}, ensuring the preservation of the original feature information:
\begin{equation}
    F_{o} = f_p^{'}(F_{class}) + F_i
\end{equation}
where ${F_{o}\in \mathbb{R}^{N\times T\times E}}$ ($o$ $\in M, K, B$) represents the output of the inner class branch.

\textbf{Inner Boundary Branch.} Our focus extends beyond action class, and we also prioritize enhancing boundary awareness for each modality.
The inner boundary branch takes ${F_{o}\in \mathbb{R}^{N\times T\times E}}$ as input, following a 1D convolutional layer $f_m(\cdot)$ to lower the feature dimension once again. 
To regress the temporal boundaries of actions, the feature dimension is decreased to $1$.
Afterward, we incorporate boundary awareness by employing several 1D convolutional layers $f_m^{'}(\cdot)$ with the residual connection.
Subsequently, the frame-level boundary probability $P_{boundary}\in \mathbb{R}^{N\times T\times 1}$ is predicted through the $Sigmoid$ activation function as
\begin{equation}
    P_{boundary} = Sigmoid(f_m^{'}(f_m(F_{o})) + f_m(F_{o}))
\end{equation}
where $P_{boundary}\in \mathbb{R}^{N\times T\times 1}$ is employed to calculate the inner boundary loss $\mathcal L_{b}^{inner}$.

In conclusion, the intermediate bottleneck module further constrains the feature extraction stage by calculating the inner class loss and the inner boundary loss.
It introduces class and boundary awareness abilities upfront for each modality, ensuring the provision of precise and valuable information for modality fusion.

\subsection{Sparse Guided Cross-Modal Module}  
\label{sec:fuse}

\begin{figure}[t]
    \centering
    \includegraphics[width= 0.75\linewidth]{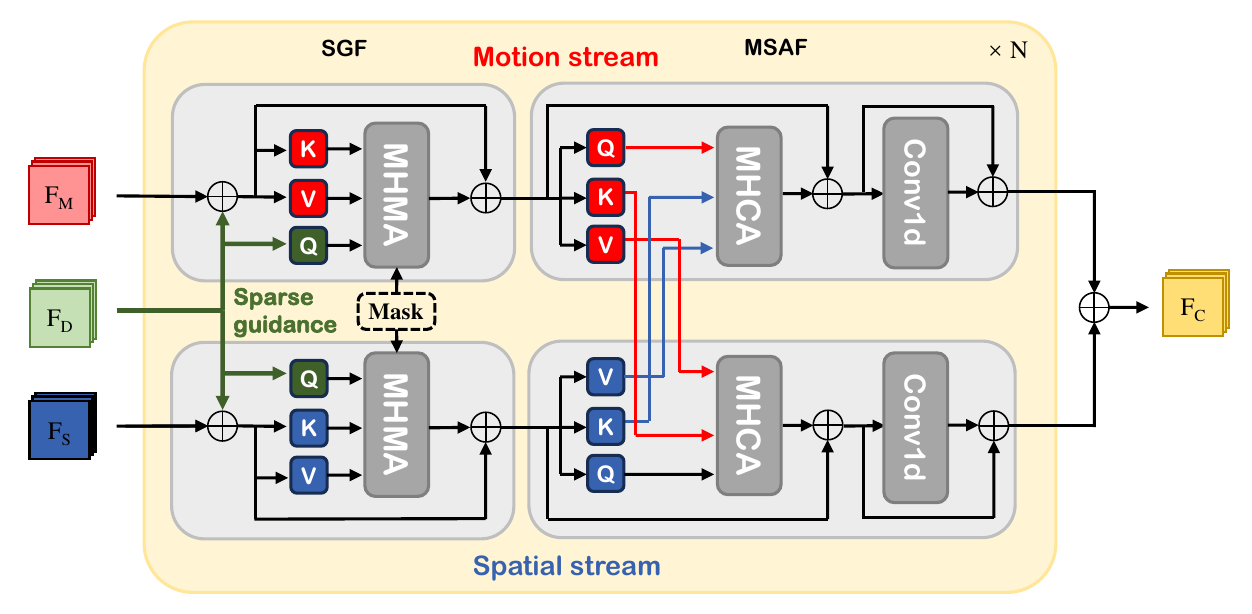}
    \caption{The architecture of the sparse guided cross-modal module. 
    It is formed by two parts: sparse guided fusion (SGF) module and motion-spatial attention fusion (MSAF) module.
    The SGF module utilizes Multi-Head Masked Attention (MHMA) to concentrate attention on temporal regions with active sparse signals, ensuring their accurate guidance of other signals.
    The Multi-Head Cross Attention (MHCA) in the MSAF is employed to capture the interaction context between motion and spatial information.
    }
    \label{fig:fusemodule}
    \Description{Fusion of four modalities utilizing motion and spatial fusion Module.}
\end{figure}
While the cross-attention in Transformers offers a straightforward approach to integrating various signals, it disregards information redundancy and emphasizes the global context, potentially leading to interference from irrelevant data. 
To address this, we employ masked attention in the sparse guided cross-modal module, which concentrates solely on feature fusion within regions where sparse signals are active, guaranteeing their precise influence on other signals.   
Moreover, IMU data captures motion details through acceleration and gyroscope information, while body key points and bounding box points offer insights into the spatial posture and boundaries of the human body. 
Consequently, we integrate multi-modal signals from both motion and spatial streams to comprehensively represent human actions.

As shown in Fig.~\ref{fig:fusemodule}, in the SGF module, we add the sparse guidance $F_D$ to the motion stream feature $F_M$ and the spatial stream feature $F_S$ to obtain the guided enhanced features, $F_{M+D}$ and $F_{S+D}$. 
Then, they are fed into a Multi-Head Mask Attention (MHMA) module together with the sparse guidance, producing the attention enhanced features $F_{M}^{SGF}$ and $F_{S}^{SGF}$ applying the following equations
\begin{equation}
    \begin{split}
        & F_{M}^{SGF} = softmax(\frac{ (\mathcal{M} \ast Q_{D} ) K_{M+D}^{T}}{\sqrt{d}})  V_{M+D}  + F_{M+D} \\
	    & F_{S}^{SGF} = softmax(\frac{ (\mathcal{M} \ast Q_{D} ) K_{S+D}^{T}}{\sqrt{d}})  V_{S+D}  + F_{S+D}.
    \end{split}
\end{equation}
Here, $d$ represents a normalization constant, %$\cdot$ represents matrix multiplication, 
$\ast$ denotes element-wise product, guided enhanced features provide keys and values for MHMA, and sparse guidance provides queries. 
Moreover, the attention mask $\mathcal{M}$ at feature location is
\begin{equation}
    \mathcal{M}^{t, e} =
        \begin{cases}
        1 & \text{ if IoT-enabled devices are used at $t$ } \\
        0 & \text{ otherwise }
        \end{cases}
\end{equation}
where $t \in [0, T]$, $e \in [0, E]$, the mask is performed on the temporal dimension, so the feature elements under the same $t$ are all $1$ or $0$.
The application of masks restricts the attention operation to instances when the sparse signals are valid. 
This precision guarantees that only accurate information from the sparse signals is integrated with the other signals, without impeding the recognition of other action categories.

Afterward, in the MSAF module, $Q_h$, $K_h$, $V_h$ $(h \in M, S)$ indicating the
query, key, and value feature of each stream are fed into the Multi-Head Cross Attention (MHCA) module. 
We utilize the query feature $Q_h$ of one stream to fetch the key feature $K_h$ and the value feature $V_h$ of the other stream as
\begin{equation}
    \begin{split}
        & F_{M \rightarrow S} = softmax(\frac{Q_{S} K_{M}^{T}}{\sqrt{d}}) V_{M}\\
	    & F_{S \rightarrow M} = softmax(\frac{Q_{M} K_{S}^{T}}{\sqrt{d}}) V_{S}
    \end{split}
\end{equation}
where $F_{M \rightarrow S}$ and $F_{S \rightarrow M}$ are the cross-stream attention features encoding the correlation between the motion stream and the spatial stream. 
Subsequently, the cross-stream attention features are merged into the fine-grained action features by frame-wise 1D convolutional layers $f_w(\cdot)$ as
\begin{equation}
    \begin{split}
        & F_{M}^{'} = f_w(F_{M \rightarrow S} + F_{M}^{SGF}) + F_{M \rightarrow S} + F_{M}^{SGF}  \\
	    & F_{S}^{'} = f_w(F_{S \rightarrow M} + F_{S}^{SGF}) + F_{S \rightarrow M} + F_{S}^{SGF}
    \end{split}
\end{equation}
where the outputs $F_{M}^{'}$ and $F_{S}^{'}$ represent the complete information interaction features of the two streams, obtained from the MSAF module. 
After multiple SGF and MSAF modules, we effectively utilize complementary information and fully integrate information from both motion and spatial streams.
Finally, the fine-grained motion feature and spatial feature are combined through an element-wise addition operation, resulting in a fused feature $F_C$ that represents human actions comprehensively.

In conclusion, the sparse guided cross-modal module delves into an in-depth analysis of human behaviors, encompassing both motion and spatial aspects. 
It incorporates sparse guidance and fosters sufficient interaction between the motion-spatial streams to obtain a comprehensive representation of the actions. 
The two streams can be extended to incorporate various heterogeneous information streams, such as RGB and skeletons, RGB and depth maps, and so on.

\subsection{Mutual Interactive Branches} 
\label{sec:fual-branch}
For the problem of over-segmentation, most previous work utilized multi-stage refinement~\cite{farha2019ms, li2020ms, yi2021asformer} and proposed boundary-aware networks~\cite{wang2020boundary, ishikawa2021alleviating}, {\itshape etc.} 
However, these studies disregard the intrinsic interplay between action class and temporal boundary, making it challenging to achieve proper trade-offs and balance during refinement. 
Instead, we explicitly model the interrelationship between action class and boundary within the mutual interactive branches. 

The two branches are each composed of multiple Transformer encoder layers. 
To expedite the acquisition of approximate boundary details, the information interaction in the initial stage is positioned at the onset of the boundary branch. 
We employ MHCA to retrieve the key feature $K_B$ and the value feature $V_B$ from the boundary branch utilizing the query feature $Q_C$ obtained from the first encoder layer of the class branch as
\begin{equation}
    F_{BD} = softmax(\frac{Q_{C} K_{B}^{T}}{\sqrt{d}}) V_{B}
\end{equation}
where $F_{BD}$ is the boundary feature after first information interaction, $d$ is a normalization constant. 
Afterward, the feature $F_{BD}$ can obtain a strong ability of temporal comprehension through the boundary branch. 

To improve the precision of frame-wise class predictions, particularly in cases where actions undergo slight changes, the second-stage information interaction occurs at the end of the two branches. 
At this stage, the boundary branch supplies the query $Q_B^{'}$, while the class branch provides the key $K_C^{'}$ and value $V_C^{'}$ as
\begin{equation}
    F_{CD} = softmax(\frac{Q_{B}^{'} {K_{C}^{'}}^{T}}{\sqrt{d}}) V_{C}^{'}
\end{equation}
where $F_{CD}$ is the class feature after the second information interaction. 

Upon reaching the end of the two branches, fully connected layers are employed to adjust the feature dimensions to the number of classes and 1, respectively.
As a result, the network produces frame-wise predictions for both action classes and temporal boundaries. 
Following that, we apply a threshold $\alpha \in [0,1]$ to filter out low-probability predictions from the confidence sequence given by the boundary branch as
\begin{equation}
    %T_B = Sigmoid(T_B^{'}) - \alpha
    B^{t}=
        \begin{cases}
        Sigmoid(B^{'t})& \text{ $ Sigmoid(B^{'t}) \geqslant \alpha $ } \\
        0 & \text{ $ Sigmoid(B^{'t}) < \alpha $ }
        \end{cases}
\end{equation}
where $B^{t}$ is the predicted temporal boundary for $t^{th}$ frame,  $B^{'t}$ is the confidence score for $t^{th}$ frame from the boundary branch.
Finally, we utilize the local barrier pooling~\cite{wang2020boundary} to refine the frame-level action classes at the predicted temporal boundaries.

In conclusion, through the mutual interactive branches, the information about action categories and temporal boundaries interact with each other. 
The results of the two branches are utilized for refinement, which ultimately leads to obtaining the frame-level refined class predictions.

\subsection{Loss Function}
In this section, we briefly introduce the loss functions employed in SigFormer.

\textbf{Frame-wise Classification Loss.}
We adopt the cross-entropy loss as a \underline{c}lassification loss:
\begin{equation}\label{eq_ce}
  \mathcal L_{c} = \frac{1}{T} \sum_{t}^{T}{-\log(p_{t,c})} 
\end{equation}
where $y_{t,c}$ is the frame-wise prediction action probability for class $c$ at time $t$. 
Features from IMU data, body key points, and body bounding box points are processed separately through the intermediate bottleneck module. 
In the inner class branch, we calculate the class loss for each feature according to Equation~\ref{eq_ce}.
Summing these three class losses yields the inner class loss, denoted as $\mathcal L_{c}^{inner}$.
Furthermore, the mutual interactive branches eventually yield frame-level refined class predictions, and the final class loss is computed utilizing Equation~\ref{eq_ce}, denoted as $\mathcal L_{c}^{final}$.

\textbf{Frame-wise Boundary Loss.}
The binary logistic regression loss function is adopted for the action \underline{b}oundary regression:
\begin{equation}\label{eq_bce}
  \mathcal L_{b} = -\frac{1}{T} \sum_{t}^{T}{(y_t\log(p_{t}) + (1-y_t)\log(1-p_t))} 
\end{equation}
where $y_t$ and $p_t$ are the ground truth and the action boundary probability for time $t$, respectively.
Similar to the inner class branch, the inner boundary branch computes boundary loss for each feature utilizing Equation~\ref{eq_bce}, summed to derive the inner boundary loss, denoted as $\mathcal L_{b}^{inner}$.
In addition, the output of the boundary branch within the mutual interactive branches is utilized to calculate the final boundary loss with Equation~\ref{eq_bce}, denoted as $\mathcal L_{b}^{final}$.

\textbf{Total Loss.}
%The SigFormer generates the inner loss ( inner class loss $\mathcal L_{c_{inner}}$ and inner boundary loss $\mathcal L_{b_{inner}}$) through the intermediate bottleneck module and the final loss ( final class loss $\mathcal L_{c_{final}}$ and final boundary loss $\mathcal L_{b_{final}}$) through the mutual interactive branches. 
\textcolor{red}{
% Features extracted from IMU data, body key points, and body bounding box points undergo separate processing through the intermediate bottleneck module. 
% Each feature is utilized to compute the class loss according to Equation~\ref{eq_ce} and the boundary loss according to Equation~\ref{eq_bce}. The three class losses are aggregated to derive the inner class loss, denoted as $\mathcal L_{c}^{inner}$, while the three boundary losses are added to derive the inner boundary loss, denoted as $\mathcal L_{b}^{inner}$. 
% The mutual interactive branches eventually yield frame-level refined class predictions, and the final class loss is computed according to Equation~\ref{eq_ce}, denoted as $\mathcal L_{c}^{final}$.
% The boundary branch within the mutual interactive branches predicts the probability that each time $t$ serves as a temporal action boundary. 
% The output of this branch is utilized to compute the final boundary loss, as defined by Equation~\ref{eq_bce}, denoted as $\mathcal L_{b}^{final}$.
% (for Reviewer2's Question1, 3 and 4)
}
Consider that action segmentation entails not only categorizing action classes but also accurately identifying temporal boundaries between actions.
Therefore, we aim to enhance each modal feature's awareness of action category and temporal action boundary by constraining the feature extraction process utilizing $\mathcal L_{c}^{inner}$ and $\mathcal L_{b}^{inner}$.
This ensures relevant information for action segmentation is extracted.
Additionally, we aim to improve the overall model's ability to predict temporal action boundaries with $\mathcal L_{b}^{final}$, ensuring that the refinement phase results in smooth and stable action class predictions.
$\mathcal L_{c}^{final}$ is utilized to optimize the final segmentation results of the model.
The four losses are combined linearly as follows:

\begin{equation}
  \mathcal L_{Total} = \lambda_1\mathcal L_{c}^{inner} + (1-\lambda_1)\mathcal L_{c}^{final} + \lambda_2(\mathcal L_{b}^{inner} + \mathcal L_{b}^{final})
\end{equation}
where $\mathcal L_{Total}$ is the total loss of the SigFormer, $\lambda_1$ is utilized to balance the inner loss and final loss, that is set to 0.5 in our experiments, and $\lambda_2$ is employed to balance the class branch and boundary branch, which is set to 0.2 in our experiments.
During training, we optimize the $\mathcal L_{Total}$ to enhance the overall performance of the model.

\section{EXPERIMENTS}
\label{sec:experiment}
\subsection{Dataset}  
\textbf{OpenPack~\cite{yoshimura2022openpack}} is a multi-modal large-scale dataset that is specifically designed for action segmentation in industrial domains. 
It contains 53.8 hours of multi-modal sensor data collected from 16 different subjects with varying levels of packaging work experience, including key points, depth images, sensory data from body-worn inertial measurement units, LiDAR point clouds, and readings from IoT-enabled devices (e.g., handheld barcode scanners utilized in work procedures). 
It consists of 20,129 instances of activities (operations) and 52,529 instances of actions with 9 different modalities. 
Since some modalities are incomplete, we select four of them that are complete and crucial in our experiments: IMU data, human key points, bounding box points, and state signals of IoT-enabled devices.

\subsection{Evaluation Metrics} 
Following ~\cite{lea2017temporal}, we analyzed segmentation performance employing the segmental F1 score. 
In particular, if more than one right detection is within the span of a single true action, 
only one is reported as a true positive, while the others are false positives. 
We calculate precision and recall for true positives, false positives, 
and false negatives across all action classes and compute the F1 score as
\begin{equation}
  F1 = 2 \times \frac{Prec\times Recall}{Prec + Recall} 
\end{equation}
The F1 score is calculated for each class and the macro average of them all is utilized to determine the measure. Following~\cite{yoshimura2022openpack}, segments corresponding to the ``Null'' category are excluded before evaluation.

\subsection{Implementation Details}

In this section, we introduce the implementation details of our experiments.

\textbf{Dataset Split.}  
For a fair comparison, 
our method is evaluated by applying a 5-fold cross-validation average for the OpenPack dataset following the previous work~\cite{tomoon}.

\textbf{Model Architecture.}
The feature extraction module consists of 6 encoder layers, while the multi-modal fusion process comprises 4 sparse guided cross-modal modules. The two mutual interactive branches adopt 3 Transformer encoder layers respectively. We employ 20 heads in both the MHA module and the MHCA module. Additionally, the boundary threshold $\alpha$ in the filter is set to 0.85.

\textbf{Training Details.}
At the feature extraction stage, we set the feature dimension of IMU and human body key points to 300, the feature dimension of body bounding box points to 240, and the feature dimension of the state signals of the scanner and printer to 60. 
We adopt the Adam~\cite{kingmaadam} optimizer with the base learning rate of $1 \times 10^{-4}$ with a $1 \times 10^{-4}$ weight decay. 
The first 50 training epochs are set as a warm-up phase and the learning rate is halved gradually after every 50 epochs. 
We exclude the ``Null'' class from the evaluation when calculating the losses. 
The model is trained for 200 epochs for each split. We utilize the batch size of 32 during training.

\subsection{Comparison with State-of-the-Art Methods}

\begin{table}[htp]
  \caption{Comparison with state-of-the-art methods on the OpenPack dataset. $I$ denotes the IMU data, $K$ denotes the human body key points, $B$ represents the human body bounding box points, $D$ denotes the state signals of IoT-enabled devices, and $+$ represents utilizing the information from multiple modalities. ``(Late)'' denotes the late fused version of the model with $I$ and $D$. ``U0104''-``U0207'' represent 6 independent test sets. Some methods have no results on ``U0203''-``U0207'' due to different dataset splits.
}
  \label{tab:all}
  \begin{tabular}{ccccccccc}
    \toprule
    \multirow{2}{*}{Modality}& \multirow{2}{*}{Methods}  &   \multicolumn{7}{c}{F1 Score (Macro Average)}  \\
    &  &    U0104    &    U0108    &    U0110    &    U0203    &    U0204    &    U0207    &    All     \\
    \midrule
    \multirow{4}{*}{$I$} & DCL-SA~\cite{singh2020deep} &  0.762& 0.782 &0.643& - & - & - &  0.752   \\
    & Conformer~\cite{kim2022inertial}  &0.801 &0.724 &0.676& - & - & -  &0.748  \\
    & LOS-Net~\cite{yoshimura2022acceleration}  &0.790 &0.793 &0.611 & - & - & -  &0.755 \\
    & \textbf{SigFormer (Ours)}  &0.866 &0.844 &0.714 & 0.847 & 0.625 & 0.701  &0.791 \\ 
    \midrule
    \multirow{3}{*}{$K$} & ST-GCN~\cite{yu2018spatio}  &0.861 &0.858 &	0.875 &	0.882 &	0.785 &	0.793 & 0.858 \\
    &MS-G3D Net~\cite{liu2020disentangling}  &0.905 &0.886 &	0.883 &	0.899 &	0.785 &	0.810 & 0.877 \\
    & \textbf{SigFormer (Ours)}  &0.913 &	0.915 &	0.885 &	0.902 &	0.789 &	0.812 & 0.889 \\
    \midrule
    \multirow{5}{*}{$I+D$} & DCL-SA (Late)~\cite{singh2020deep} & 0.820	& 0.807	& 0.725	& -	&-	&-	&0.797	\\
    & Conformer~\cite{kim2022inertial} &0.834&	0.778	&    0.716	&-	&-	&-	&0.787	\\
    & LOS-Net (Late)~\cite{yoshimura2022acceleration}&0.816	&0.813	&0.644	&-	&-	&-	&0.777	\\
    & LTS-Net~\cite{yoshimura2022openpack} &0.847	&0.842	&0.757	&-	&-	&-	&0.830	\\
    & \textbf{SigFormer (Ours)}  &0.852 &0.830 &0.881 & 0.858 & 0.752 & 0.808  &0.844 \\ 
    \midrule
    \multirow{2}{*}{$I+K$} & Chai {\itshape et al.}~\cite{openpack3}&0.954&0.932&0.912&0.948&0.831&0.813	&0.917	\\
    & \textbf{SigFormer (Ours)}  &0.967 &0.962 &0.911 & 0.961 & 0.831 & 0.913  &0.937 \\ 
    \midrule
    \multirow{2}{*}{$I+K+B+D$} & Uchiyama~\cite{tomoon} &0.960	&0.956	&0.912	&0.955	&0.839	&0.885	&0.932	\\
    & \textbf{SigFormer (Ours)} &\textbf{0.971}	&\textbf{0.969}	&\textbf{0.960}	&\textbf{0.966}	&\textbf{0.903}	&\textbf{0.923}	&\textbf{0.958}	\\
  \bottomrule
\end{tabular}
\end{table}

\begin{figure}[t]
    \centering
    \includegraphics[width=\linewidth]{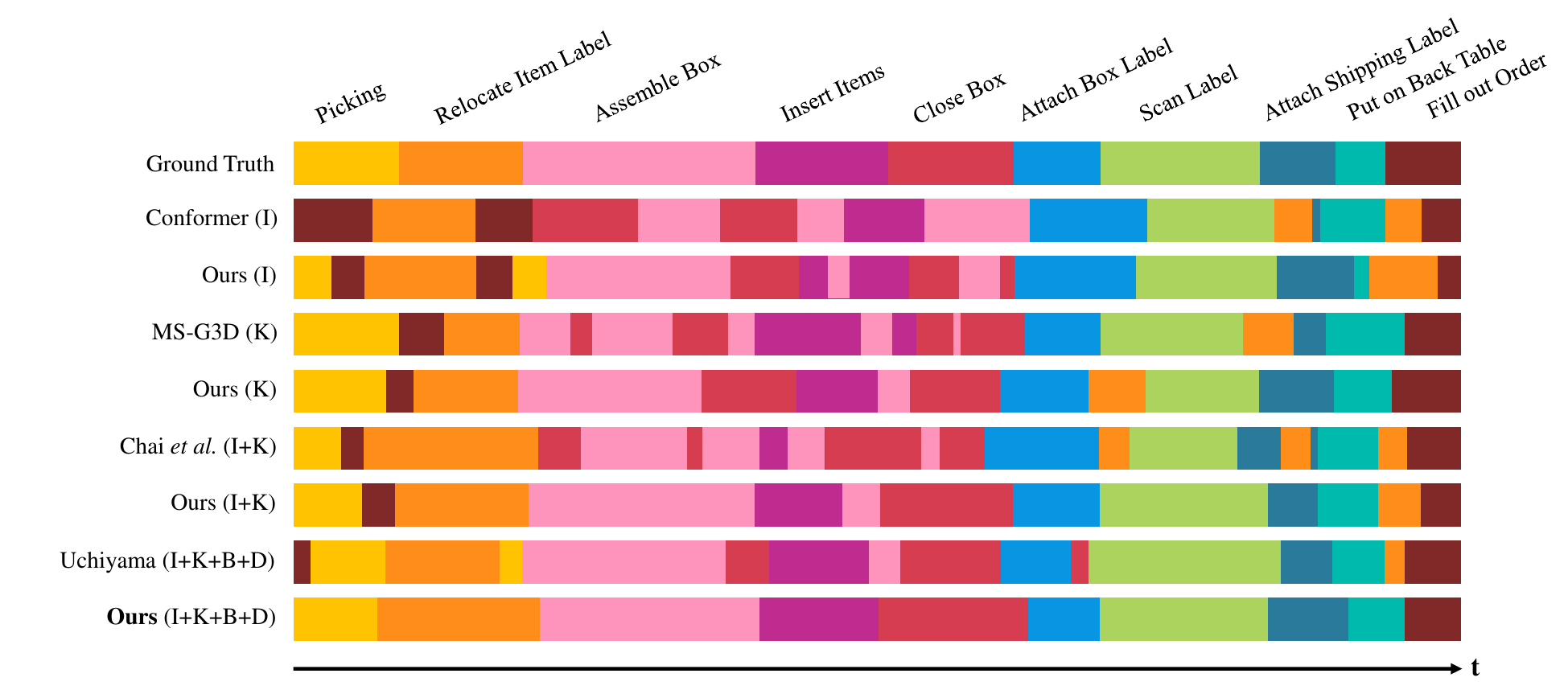}
    \caption{Qualitative results on OpenPack. SigFormer effectively utilizes multi-modal information and alleviates the issue of over-segmentation more efficiently compared to other methods.}
    \label{fig:resultopenpack}
    \Description{Qualitative results on OpenPack. It continues our method and other baseline methods. It shows that our method can exploit multi-modal information better than other methods and solve the problem of over-segmentation.}
\end{figure}

As illustrated in Table~\ref{tab:all}, we compare our SigFormer with previous state-of-the-art methods on the OpenPack dataset, including IMU-based methods such as  DCL-SA~\cite{singh2020deep}, Conformer~\cite{kim2022inertial}, LOS-Net~\cite{yoshimura2022acceleration}, skeleton-based methods such as ST-GCN~\cite{yu2018spatio}, MS-G3D Net~\cite{liu2020disentangling}, multi-modality methods such as LTS-Net~\cite{yoshimura2022openpack}, methods of Chai {\itshape et al.}~\cite{openpack3} and Uchiyama~\cite{tomoon}. In addition, models that incorporate state signals of IoT-enabled devices in IMU data \cite{singh2020deep, kim2022inertial, yoshimura2022acceleration} are also introduced for comparison as multi-modal models.
To ensure a fair comparison, we evaluate SigFormer under the same combinations of modalities. 

First, we compare SigFormer with other methods in the case of single-modality input.
SigFormer gets an F1 score of 0.791 based on IMU data, surpassing LOS-Net~\cite{yoshimura2022acceleration} by 3.6\%. 
Additionally, it outperforms MS-G3D Net~\cite{liu2020disentangling} by 1.2\% when utilizing human key points data. 
The excellent performance in single-modality action segmentation proves the efficacy of SigFormer in capturing the intrinsic characteristics of individual data sources.

Secondly, we compare SigFormer with other methods in the context of multi-modality inputs.
SigFormer obtains an F1 score of 0.844 with the combination of ``I+D'', demonstrating a 1.4\% improvement over LTS-Net~\cite{yoshimura2022openpack}.
Furthermore, SigFormer achieves an F1 score of 0.937 based on ``I+K'', which is 2.0\% higher than the method of Chai {\itshape et al.}~\cite{openpack3}.
Notably, Uchiyama~\cite{tomoon} integrated models trained through five different methods. 
When employing the ``I+K+B+D'' modality combination, their first training method got an F1 score of 0.932. 
Adhering to identical training configurations, including modality types and data processing methodologies, SigFormer obtains the macro-average F1 score of 0.958, exhibiting a significant improvement of 2.6\% compared to their method. 

Especially on the U0110, U0204, and U0207 test sets, compared to Uchiyama~\cite{tomoon}, SigFormer demonstrates significant performance improvements, with an average enhancement of 5\%.
% These test sets primarily comprise short-duration actions, posing challenges for action segmentation. 
The average duration of actions in these test sets ranges from 7 to 8 seconds, whereas in other test sets, it spans from 10 to 13 seconds. 
Shorter action durations pose a more significant challenge for action segmentation.
SigFormer incorporates action boundary information as constraints in both the feature extraction and final prediction phases. 
This approach enables accurate identification of temporal action boundaries, thereby enhancing the model's ability to predict short action segments.

Moreover, SigFormer achieves the F1 score of over 0.9 on all six independent test sets, indicating its robustness for action segmentation.
Under each of the same modality combinations, SigFormer exhibits superior performance. 
Even utilizing fewer modalities (``I+K''), SigFormer attains better results than Uchiyama~\cite{tomoon}, which requires more modalities as input.

Fig.~\ref{fig:resultopenpack} visualizes the action segmentation results of several state-of-the-art models on OpenPack to quantify the impact of our method on multi-modal fusion and eliminating over-segmentation errors. 
We can observe that our method achieves better action class segment prediction when utilizing the same modality, demonstrating enhanced modeling of inter-modal relationships. 
Furthermore, the methods from Uchiyama~\cite{tomoon} and Chai {\itshape et al.}~\cite{openpack3} suffer from over-segmentation errors, whereas our approach produces segmentation results that avoid the occurrence of small incorrect segments at action boundaries.

As a result, SigFormer achieves an impressive macro-averaged F1 score of 0.958 across all test sets, surpassing the performance of the current state-of-the-art methods utilizing the same modality.

\subsection{Ablation Study}
In this section, we introduce the ablation studies to evaluate the effectiveness of each module in SigFormer on the OpenPack dataset. 

\begin{table}
  \caption{The experiments on different combinations of modality.
  }
  \label{tab:ab1}
  \begin{tabular}{cccccc}
    \toprule
    \multicolumn{4}{c}{Modality} &  \multirow{2}{*}{F1 (Macro Average)} \\
    $I$ & $K$ & $B$ & $D$ & &\\
    \midrule
    \checkmark   &	&	&		&0.791 \\
     &	\checkmark	&	&		&0.889\\
      &	&\checkmark	&		&0.833\\
    & & &\checkmark   &0.191\\
    \midrule
    \checkmark	&\checkmark	&	&		&0.937\\
    &	\checkmark	&\checkmark	&		&0.921\\
    \checkmark	&	&\checkmark	&		&0.918\\
    
    \checkmark	&\checkmark	&	&\checkmark		&0.947\\
    \checkmark	&\checkmark	&\checkmark	&		&0.950\\
    \checkmark	&\checkmark	&\checkmark	&\checkmark	 &\textbf{0.958}\\
  \bottomrule
\end{tabular}
\end{table}

\begin{figure}[t]
    \centering
    \includegraphics[width=\linewidth]{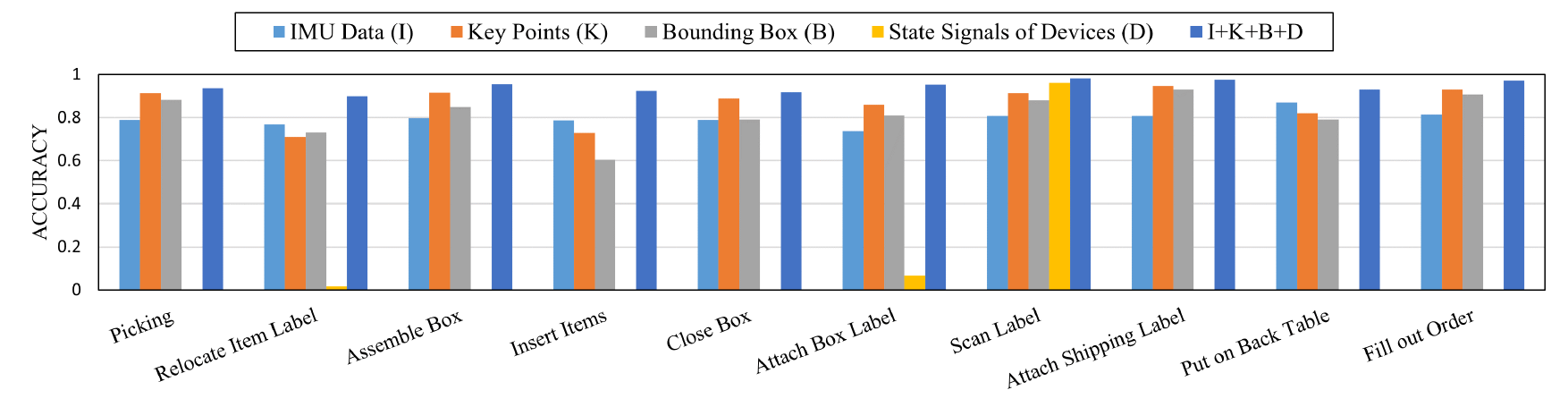}
    \caption{Accuracy on each action class for each modality and combination of them.}
    \label{fig:eachmodality}
    \Description{Accuracy on each action class for each modality and combination of them. Human key points data performed best, followed by bounding box data, and IMU data performed worst. The state signals of IoT devices provide precise but sparse action cues.}
\end{figure}

\subsubsection{Single-Modal Action Segmentation.} 
To investigate the characteristics of each modality, we present the accuracy in action segmentation on each action class for each modality in Fig.~\ref{fig:eachmodality}. 
Among these modalities, 
We discover that human key points usually dominate the performance and achieve the highest accuracy on the majority of action classes, which is consistent with the results in Table~\ref{tab:ab1}. 
The reason may be that the human key points supply a wealth of spatial information, characterize the details of the relative position changes of human joints, and improve the discrimination of actions. 
The body bounding box points simply represent the person's general location in space, and the spatial details are fewer than the human key points, resulting in poor performance. 
With motion information, IMU data is effective in distinguishing high-acceleration actions, such as ``insert'' and ``put''. 
IMU data have the lowest performance (excluding state signals of devices), possibly because they only carry motion information and are unable to handle changes in spatial location and human-object interactions. 
On the other hand, state signals of IoT-enabled devices, i.e., the handheld scanner and label printer, are highly trustworthy sources of information for recognizing specific operations.
As shown in Figure~\ref{fig:eachmodality}, these signals demonstrate the highest accuracy in recognizing actions that involve the corresponding devices, such as "scanning label". 
However, they are less effective in recognizing most other actions. Consequently, as indicated in Table~\ref{tab:ab1}, the F1 score for predicting all actions solely utilizing state signals of the handheld scanner and label printer is only 0.191.
%As a result, these signals exhibit the highest accuracy in recognizing actions that have utilized the corresponding devices, such as ``scanning label''. 
%However, they are not effective in recognizing the majority of other actions.

\subsubsection{Multi-Modal Action Segmentation.} We try different combinations of multiple modalities and summarize the F1 scores in Table~\ref{tab:ab1}. 
%Intuitively, the more modalities are involved, the better performance we obtain, since complementary information can be utilized to compensate missing features in the single modal data. 
Intuitively, the more modalities involved, the better performance we obtain, as complementary information can compensate for missing features in single-modal data.

First, we discuss the case of two modalities as input.
By fusing the wrist motion features from the IMU data with the skeleton spatial features extracted from the key points, the F1 score reaches 0.937.
Since the human bounding box data only provides an approximate location of the entire human body in space, it contains less spatial information compared to the key points.
Consequently, the F1 score of the ``I+B'' combination is lower than that of ``I+K''.
When the two modalities that solely reflect spatial information are fused (``K+B''), the absence of motion information from the IMU data limits the perspective of understanding human actions, resulting in a relatively lower F1 score of 0.921.

Next, we discuss the case of the fusions of three modalities.
When incorporating body bounding box data in addition to IMU data and key points, the human position information is further integrated, resulting in a 1.4\% increase in the F1 score.
Additionally, the introduction of state signals from IoT-enabled devices brings an improvement of roughly 1.0\% to the F1 score.

Finally, by fusing all four modalities, the model achieved the highest F1 score of 0.958.
As shown in Fig.~\ref{fig:eachmodality}, the combination of all different modalities enables the complementarity of information between them, thereby obtaining an accurate representation of human actions.
 
In conclusion, the fusion of IMU data, human body key points data, and body bounding box data effectively captures the comprehensive action information, encompassing both the motion stream and spatial stream. 
Furthermore, incorporating information from IoT-enabled devices further enhances the model's performance, providing precise action cues.

\begin{table}[t]
  \caption{Comparisons of different feature fusion strategies on the OpenPack dataset. }
  \label{tab:ab2}
  \begin{tabular}{cc}
    \toprule
    Fusion Strategy  & {F1 (Macro Average)} \\
    \midrule
    Add~\cite{openpack3} & 0.941 \\
    Concatenate~\cite{tomoon} &0.942 \\
    BF blocks~\cite{yoshimura2022openpack} & 0.944 \\
    MCFM~\cite{inoshita2023exploring} & 	0.946 \\
    TRN~\cite{zhou2018temporal} & 0.937 \\
    \midrule
    Only MSAF  & 0.947 \\ 
    Only SGF  & 0.945 \\
    Cross-attention + MSAF & 0.949 \\
    \textbf{SGF + MSAF (Ours)} & \textbf{0.958} \\
  \bottomrule
\end{tabular}
\end{table}

\begin{figure}[t]
    \centering
    \includegraphics[width=\linewidth]{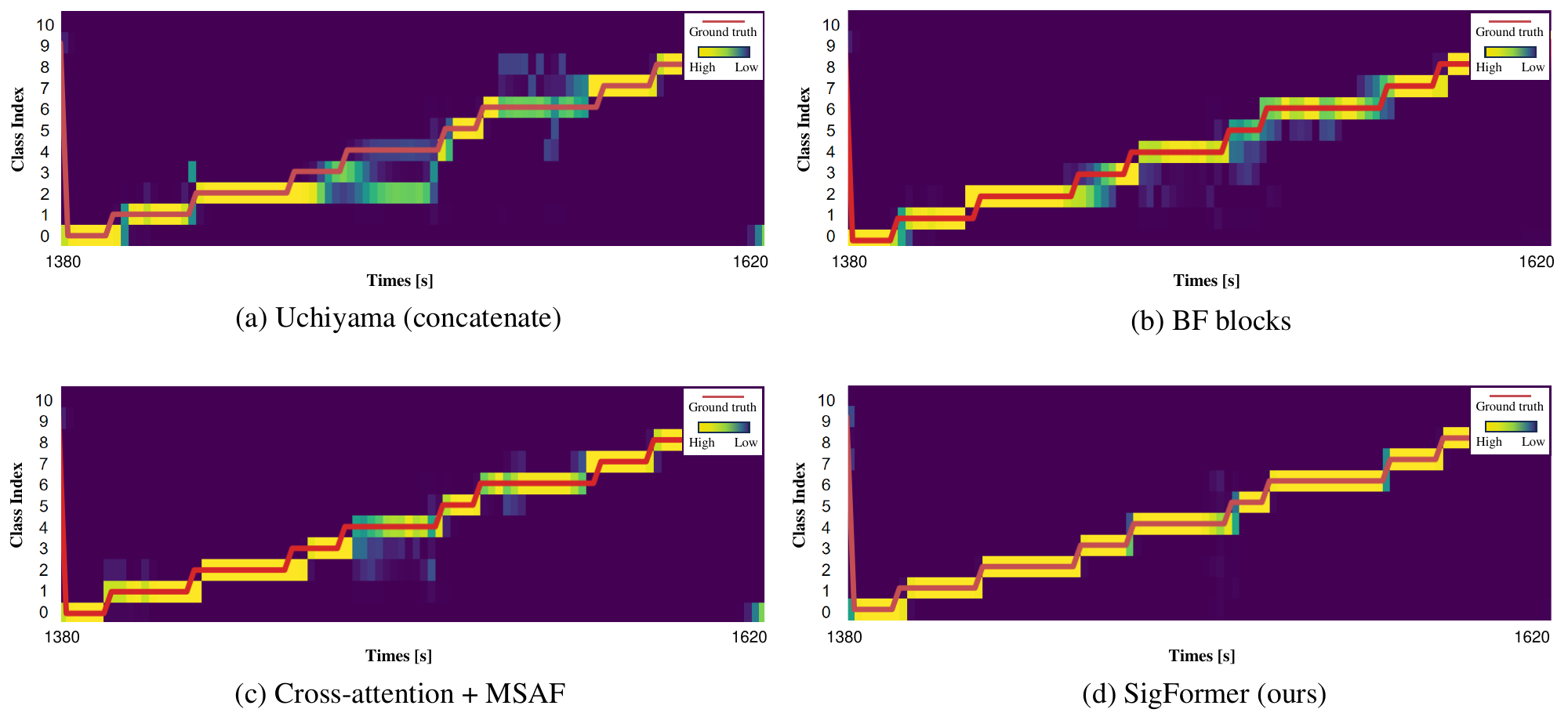}
    \caption{The prediction results obtained from models employing various fusion methods. (a) Concatenate multi-modal features. 
    (b) BF blocks leverage the high-confidence sparse signals as anchors and fuse them with dense sensor signals.
    (c) The mask attention in the SGF module is replaced by normal cross-attention to demonstrate its effectiveness.
    (d) Ours.
    The figures display the time-series data of the ground truth (red line), along with the class estimates represented as probabilities. High- and low-class probabilities are yellow and blue, respectively.
    SigFormer's predictions closely align with the ground truth, effectively distinguishing similar actions and avoiding unstable estimates.
    }
    \label{fig:visual}
    \Description{The visualization of models' prediction results on OpenPack.}
\end{figure}

\subsubsection{The Assessment of Different Feature Fusion Approaches.} We conducted experiments on various feature fusion methods shown in Table~\ref{tab:ab2}. 
The F1 score for the naive feature fusion strategies~\cite{openpack3, tomoon} is around 0.943. 
The strategies of BF blocks~\cite{yoshimura2022openpack} and MCFM~\cite{inoshita2023exploring} improve slightly, but still fall short in accurately capturing the relationship between multi-modality features. 
It is worth noting that TRN~\cite{zhou2018temporal} for multi-modality fusion resulted in a significant decrease in the F1 score compared to other methods. 
This could be because non-critical information from each modality is repeatedly combined, interfering with the representation of the actions. 
 
When solely relying on the MSAF module, the sparse signals do not guide the multi-modal signals, resulting in suboptimal recognition of specific action categories. 
Conversely, when employing only the SGF module, the F1 score reaches 0.945, demonstrating that fusion performance remains moderate due to insufficient interactivity among the signals, even though they are accurately guided by the sparse signal. 
Notably, we substituted the SGF module with a standard cross-attention operation to validate the significance of the masked attention within it.

As shown in Fig.~\ref{fig:visual}, we visualize the predictions of different fusion methods at the $softmax$ layer.
The straightforward concatenation method exhibits category prediction errors over extended periods, indicating its limited utilization of complementary information from various modalities. While the predictions from BF blocks and standard cross-attention are generally accurate, instances of category confusion and instability still arise.
Compared to other methods, our proposed sparse guided cross-modal module reduces the potential confusion between similar action categories.
Of particular note is that SigFormer exhibits increased stability in action category predictions linked to sparse signals, specifically those categorized under class index 6 (``scanning label'').
This observation underscores SigFormer's efficacy in integrating multi-modal signals characterized by varying information densities to yield precise representations of human actions.

\begin{table}
  \caption{Ablation study of adding the intermediate bottleneck module after the feature extraction of different modalities. To enhance the clarity of the results, we experimented with inner class loss in a network setting that excluded the inner boundary branch and the final boundary branch.}
  \label{tab:ab3}
  \begin{tabular}{cccccc}
    \toprule
    \multicolumn{4}{c}{Modality}  & \multicolumn{2}{c}{F1 (Macro Average)}  \\
    $I$ & $K$ & $B$ & $D$  & Inner Class & Inner Boundary\\
    \midrule
        &   &   &  & 0.914 & 0.938\\
        \checkmark &\checkmark  &  &  & 0.932 & 0.952\\
         \checkmark &\checkmark&\checkmark &  & \textbf{0.936} & \textbf{0.958}\\
         \checkmark &\checkmark&\checkmark&\checkmark & 0.935 & 0.950\\
  \bottomrule
\end{tabular}
\end{table}

\begin{table}
  \caption{Comparisons of different loss choices for SigFormer.}
  \label{tab:abloss}
  \begin{tabular}{cccccc}
    \toprule
    \multicolumn{4}{c}{Losses}  & \multirow{2}{*}{F1 (Macro Average)} \\
    $\mathcal L_{c}^{final}$ & $\mathcal L_{b}^{final}$ & $\mathcal L_{c}^{inner}$ & $\mathcal L_{b}^{inner}$  &  \\
    \midrule
        \checkmark & & & & 0.914\\
        \checkmark & &\checkmark& & 0.936\\
        \checkmark &\checkmark  &  &  & 0.927\\
         \checkmark &\checkmark&\checkmark& & 0.938\\
         \checkmark &\checkmark&\checkmark&\checkmark & \textbf{0.958}\\
  \bottomrule
\end{tabular}
\end{table}
\subsubsection{The Effectiveness of Intermediate Bottleneck Module.} To evaluate the effectiveness of the intermediate bottleneck module, we conducted ablation experiments on the inner class loss and the inner boundary loss, and the results are shown in Table~\ref{tab:ab3}. In addition, Table~\ref{tab:abloss} compares the results of each combination of loss functions.

\textbf{Inner Class Loss.} As illustrated in Table~\ref{tab:ab3}, the performance of our method can be significantly improved by calculating the inner class loss for the three types of information-rich data (IMU data, human body key points, and body bounding box points). 
This demonstrates that the inner class loss introduces the enhanced awareness of action class for each modality, enabling effective learning of discriminative feature representations. 
However, computing inner class loss for state signals of IoT devices fails to enhance the performance.
This is probable because the state signals are relatively temporally sparse and only correspond to specific actions, making it impossible to directly determine the whole action classes from this information alone.
Moreover, as shown in Table~\ref{tab:abloss}, when inner class loss is included in the modality combinations, the model performs significantly better, which also illustrates the effectiveness of introducing class awareness for each modality.

\textbf{Inner Boundary Loss.} Similarly to the experiments of inner class loss, computing the inner boundary loss for the first three modalities produces the best performance for action segmentation (Table~\ref{tab:ab3}).
This favorable outcome can be attributed to the fact that each feature individually learns temporal boundary information, thus providing accurate boundary awareness for subsequent action segmentation. 
%In contrast, due to the sparse and indistinct temporal boundaries in the devices' state signals information, calculating the inner boundary loss for it lowers the F1 score by 0.8\%.
In contrast, due to the temporal sparsity of the device state signals, the action boundaries inferred from them are also sparse.
Consequently, calculating the inner boundary loss based on these signals introduces a bias in the model's perception of the correct action boundary,  resulting in a performance degradation of 0.8\%.
Furthermore, as demonstrated in Table~\ref{tab:abloss}, the introduction of inner boundary loss can improve the F1 score by 1.1\%, underscoring the efficacy of integrating boundary awareness within each dense signal.

In summary, incorporating the inner bottleneck module in the network and calculating the inner losses can constrain the network with relevant action class and boundary information during the feature extraction stage. This approach enables the targeted extraction of accurate and beneficial features from various modal data, thereby positively influencing subsequent tasks.

\begin{table}
  \caption{Ablation experiments on multi-stage interactive and the boundary loss weight $\lambda_2$. DB is the class-boundary dual branch. When the boundary loss weight $\lambda_2$ is set to 0, it signifies the absence of the boundary branch in the model.}
  \label{tab:ab5}
  \begin{tabular}{ccccccccc}
    \toprule
    \multirow{2}{*}{Methods}  & \multicolumn{8}{c}{Boundary Loss Weight $\lambda_2$} \\
     & 2 & 1 & 0.5 &0.4 & 0.2 & 0.1 & 0.08 & 0 \\
    \midrule
    DB w/o Multi-stage interactive   &0.918 &0.938 &0.930 &0.942 &0.942 &0.940 &0.938 &0.936 \\
    DB w/o First-stage interactive   &0.922 &0.942 &0.933 &0.948 &0.951 &0.945 &0.941 &0.936 \\
    DB w/o Second-stage interactive   &0.923 &0.941 &0.934 &0.945 &0.950 &0.949 &0.940 &0.936 \\
    DB w Multi-stage interactive    &0.922 &0.944 &0.935 &0.952  &\textbf{0.958}&0.950 &0.948 &0.936 \\
  \bottomrule
\end{tabular}
\end{table}

\begin{figure}[t]
    \centering
    \includegraphics[width=\linewidth]{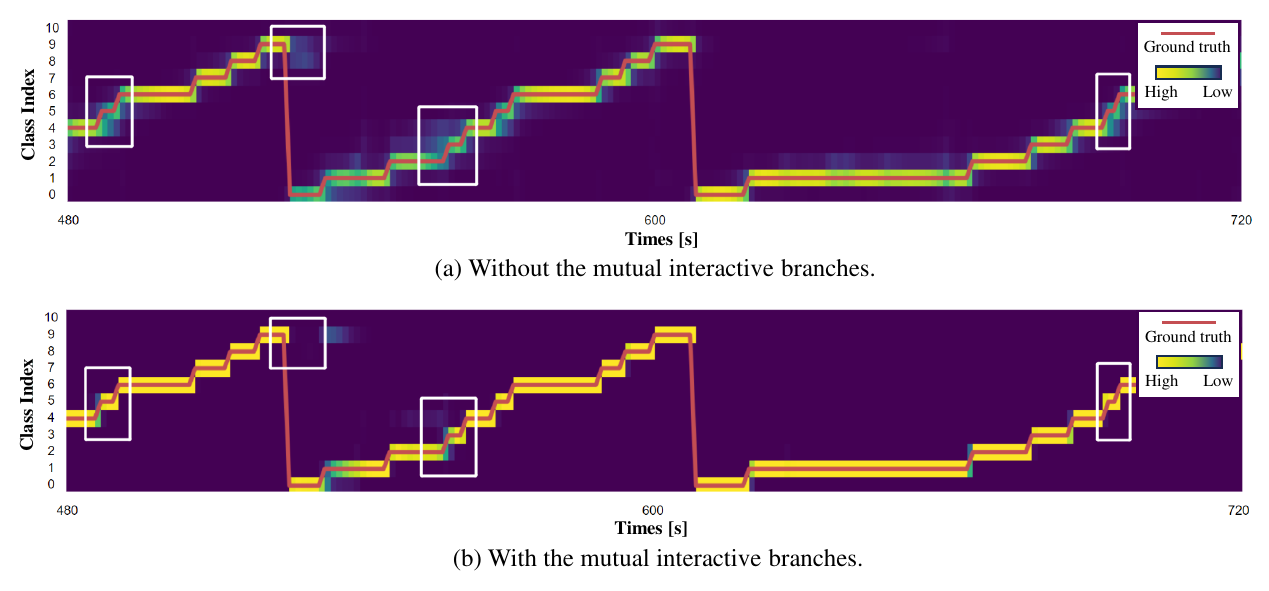}
    \caption{The prediction results obtained from models (a) without and (b) with the mutual interactive branches. 
    The figures display the time-series data of the ground truth (red line), along with the class estimates represented as probabilities. 
    High- and low-class probabilities are yellow and blue, respectively. 
    The white boxes highlight the disparities between the results of different methods.
    The mutual interactive branches notably enhance the model's prediction accuracy, particularly at temporal action boundaries.
    }
    \label{fig:visual2}
    \Description{The visualization of models' prediction results on OpenPack.}
\end{figure}

\subsubsection{The Effectiveness of Mutual Interactive Branches.} As illustrated in Table~\ref{tab:ab5}, the explicit modeling of the context of action classes and boundaries leads to an improvement of nearly 1.0\% over the results without any information interaction. 
In addition, information interaction at each stage between two branches is crucial, and its absence at any stage results in a decrease in the F1 score.

Fig.~\ref{fig:visual2} demonstrates that the two branches equipped with multi-stage information interaction produce more accurate predictions precisely at the frames where the action class transitions occur. 
This observation validates the effectiveness of modeling the context of the class branch and boundary branch.

\subsubsection{The Effectiveness of Boundary Branch Weight.}
As shown in Table~\ref{tab:ab5}, the boundary branch weight $\lambda_2$ plays a key role in balancing action classes and boundaries. 
When $\lambda_2$ is large, the model's performance is even worse than that without the boundary branch. 
This is because the model overly prioritizes temporal boundary information, disregarding the recognition of action categories. 
As $\lambda_2$ gradually decreases to 0.2, the model achieves the highest F1 score. 
This value represents a balance between the boundary loss and class loss, allowing the model to take into account both action classification and temporal boundary regression. 
However, further decreasing $\lambda_2$ below 0.2 leads to a gradual decline in segmentation performance, indicating inadequate attention to boundary information and inadequate constraints on frames when the action changes slightly.

\begin{table}
  \caption{Ablation experiments on the boundary threshold $\alpha$ in the filter.}
  \label{tab:ab6}
  \begin{tabular}{ccccccc}
    \toprule
    {Threshold~$\alpha$ }& 0.95 & 0.90 & {0.85}  & 0.80 & 0.75 & 0.70 \\
    \midrule
    {F1 (Macro Average)}	&0.921 &0.938 	&\textbf{0.958} &0.947 &0.941 & 0.930\\
  \bottomrule
\end{tabular}
\end{table}

\subsubsection{The Investigation of the Optimal Boundary Threshold.} We also explore the effectiveness of the boundary threshold. As shown in Table~\ref{tab:ab6}, the model's performance is influenced by the threshold $\alpha$ in the filter that precedes the refinement process. 
When $\alpha$ is higher, fewer boundaries are involved in the refinement process, resulting in the network smoothing the action classes at the original temporal boundaries, leading to the under-segmentation issue. 
Conversely, decreasing $\alpha$ includes non-true boundaries in the class constraints, introducing additional noise and bias that leads to over-segmentation. 
The value of $\alpha$ is dataset-dependent which can be chosen empirically utilizing a validation set. 
Our framework exhibits the greatest performance when $\alpha$ is set to 0.85.

\section{CONCLUSION}  
\label{sec:conclusion}
In this paper,  
we introduce SigFormer, an efficient network for the action segmentation task utilizing both sparse and dense signals.  
The core component of SigFormer is the sparse guided cross-modal module, which adopts the sparse signals as the guidance to fuse signals with different information densities. 
It employs masked attention to constrain the attention operation on specific temporal regions. 
Moreover, to augment the network's awareness of temporal boundaries, we develop the intermediate bottleneck module equipped with the inner loss functions, which extracts features related to both action category and temporal boundary. 
Furthermore, the mutual interactive branches are designed to explicitly model multi-stage interactions between action category and temporal boundary. 
In summary, SigFormer not only leads to improved fusion performance but also mitigates the over-segmentation errors. 
Comprehensive experiments demonstrate the superior performance of SigFormer on the OpenPack dataset by achieving a remarkable F1 score. 

\section{ACKNOWLEDGMENTS}
This work was sponsored by Beijing Nova Program (NO. 20220484063), XDC02050200.

% \begin{acks}

% \end{acks}

\bibliographystyle{ACM-Reference-Format}
\bibliography{reference}


\end{document}